\documentclass[10pt,conference]{IEEEtran}
\IEEEoverridecommandlockouts

  \usepackage{cite}
\usepackage{listings}
\usepackage{enumitem}
\usepackage{stfloats}
\usepackage{multirow}
\usepackage[pdftex]{graphicx}

\usepackage{lipsum}
\usepackage[ruled]{algorithm2e}
\usepackage{amsmath}
\usepackage{acronym}
\usepackage{algorithmic}
\usepackage{array}
\usepackage{mdwmath}
\usepackage{mdwtab}
\usepackage{eqparbox}

\DeclareMathOperator*{\sign}{sign}

\usepackage{xcolor}
\def\BibTeX{{\rm B\kern-.05em{\sc i\kern-.025em b}\kern-.08em
        T\kern-.1667em\lower.7ex\hbox{E}\kern-.125emX}}
\begin{document}
%

\title{Conditionally Deep Hybrid Neural Networks \\Across Edge and Cloud\\
    \thanks{The research was funded in part by C-BRIC, one of six centers in JUMP, a Semiconductor Research Corporation (SRC) program sponsored by DARPA, the National Science Foundation, Intel Corporation, Vannevar Bush Faculty Fellowship, and the U.K. Ministry of Defence under Agreement Number W911NF-16-3-0001.}
    \vspace{-1ex}
}
\author{\IEEEauthorblockN{ Yinghan~Long\textsuperscript *, Indranil Chakraborty\textsuperscript *, Kaushik Roy }
        \textit{(* Authors contributed equally to the work)}
\IEEEauthorblockA{\textit{School of Electrical and Computer Engineering, Purdue University} \\
    long273@purdue.edu, ichakra@purdue.edu, kaushik@purdue.edu}
\\[-5.0ex]
}

%
%


\maketitle

\begin{abstract}
The pervasiveness of ``Internet-of-Things" in our daily life has led to a recent surge in fog computing, encompassing a collaboration of cloud computing and edge intelligence. To that effect, deep learning has been a major driving force towards enabling such intelligent systems. However, growing model sizes in deep learning pose a significant challenge towards deployment in resource-constrained edge devices. Moreover, in a distributed intelligence environment, efficient workload distribution is necessary between edge and cloud systems. To address these challenges, we propose a conditionally deep hybrid neural network for enabling AI-based fog computing. The proposed network can be deployed in a distributed manner, consisting of quantized layers and early exits at the edge and full-precision layers on the cloud. During inference, if an early exit has high confidence in the classification results, it would allow samples to exit at the edge, and the deeper layers on the cloud are activated conditionally, which can lead to improved energy efficiency and inference latency. We perform an extensive design space exploration with the goal of minimizing energy consumption at the edge while achieving state of the art classification accuracies on image classification tasks. We show that with binarized layers at the edge, the proposed conditional hybrid network can process 65\% of inferences at the edge, leading to 5.5$\times$ computational energy reduction with minimal accuracy degradation on CIFAR-10 dataset. For the more complex dataset CIFAR-100, we observe that the proposed network with 4-bit quantization at the edge achieves 52\% early classification at the edge with 4.8$\times$ energy reduction. The analysis gives us insights on designing efficient hybrid networks which achieve significantly higher energy efficiency than full-precision networks for edge-cloud based distributed intelligence systems.
\end{abstract}


\section{Introduction}
The age of ``Internet-of-Things" (IoT) has touched human lives in an unprecedented manner by endowing us with remarkable connectivity and autonomous intelligent systems \cite{fog_iot}. These benefits come with the necessity of processing humongous amounts of heterogeneous data obtained from the environment. The ability of deep learning to reliably process such data has resulted in it playing a dominant role in a wide range of Artificial Intelligence (AI) applications, including image classification \cite{krizhevsky2012imagenet,szegedy2015going,he2016deep}, natural language processing \cite{mikolov2013distributed}, and object detection \cite{girshick2015fast}. As the third wave of artificial intelligence (AI) is accelerating, researchers are incorporating AI into IoT applications. Traditionally, these cognitive applications perform compute-intensive tasks in a centralized hub which collects data from thousands of connected edge devices. However, in the recent past, there have been rising concerns of data security due to over-centralization of information, and the continuous communication overhead between the edge devices and the cloud drastically increases the power consumption of world-wide internet. Moreover, real-time AI systems such as health monitoring \cite{health} and autonomous driving \cite{autonomous} require fast processing. Hence, to truly realize the potential of AI in IoT, it is necessary to enable intelligence at the edge \cite{IoT_in_edge}. 

\begin{figure*}
\centering
  \includegraphics[width=5.6 in]{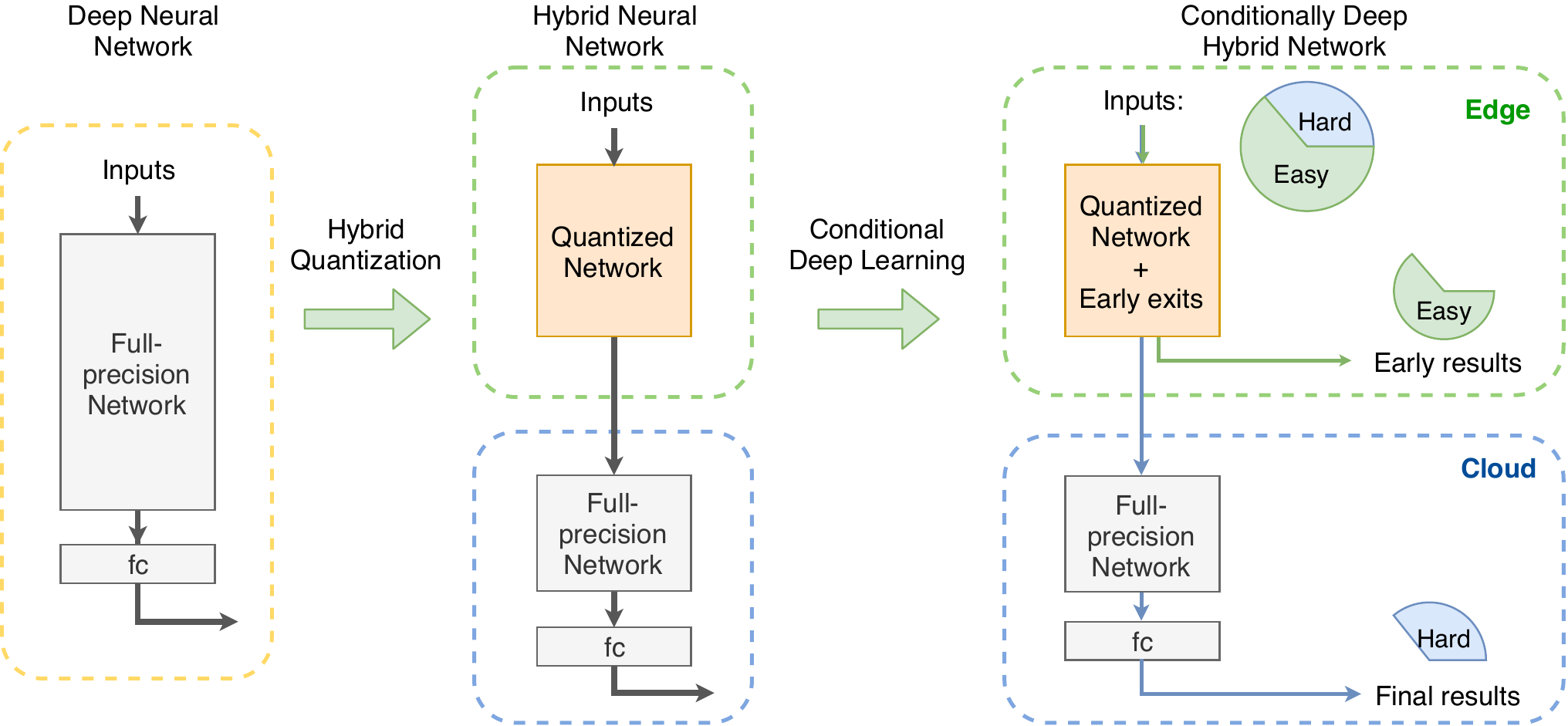}
  \caption{Overview of our design methodology for Conditionally Deep Hybrid Networks. From left to right, we show how a standard deep neural network is developed into a Conditionally Deep Hybrid Network and distributed to an edge-cloud system.}
\label{fig_method}
\end{figure*}

On the other hand, the resource constraints of edge devices limit us from deploying large deep learning models at the edge. With the growing complexity of tasks, there has been a considerable increase in the sizes of deep neural networks (DNNs). This requires DNNs to be trained on powerful Graphic Processing Units (GPUs) or Tensor Processing Units (TPUs) \cite{jouppi2017datacenter}. Edge devices, however, are resource-constrained and implementing DNNs on such devices jeopardizes the requirement of long battery life for stable operation. This has propelled the search for techniques that enable energy-efficient implementations of deep learning models. Besides, researchers are investigating efficient techniques to enable distributed intelligence at the near-user edge device and end-user cloud \cite{distributed}.

 A popular approach to reduce model complexity and improve computational efficiency is to quantize the weights and activations of neural networks by representing 32-bit floating-point values with lower bit-depth fixed point \cite{Quantization:journals/corr/abs-1712-05877} or even binary values\cite{BinaryNN:journals/corr/CourbariauxB16} \cite{XNOR:journals/corr/RastegariORF16}. Compared to full-precision networks, binary neural networks are fast and efficient, but this comes with a degradation in performance. For example, binarizing both inputs and weights of a residual network(ResNet) causes 18\% reduction in top-1 accuracy on ImageNet and 9\% on CIFAR10. In order to overcome this challenge, there have been various approaches towards improving quantization schemes \cite{zhou2016dorefa, zhou2017balanced} as well as proposing training algorithms for networks with increased bit-precision \cite{zhang2018lq, jung2018joint, choi2018pact}. An alternative approach towards preserving the performance while achieving a significant energy-efficiency is designing hybrid network architectures \cite{chakraborty2019pca, prabhu2018hybrid}, which consist of both quantized and high-precision layers. Since high-precision layers are power-hungry, such computations can be performed in the cloud. 

In this work, we propose a design methodology to build conditionally deep hybrid networks for distributed intelligence across an edge-cloud system. In such a network, quantized layers of the network are deployed in the edge device, while high-precision layers are performed on the cloud. Further, by using the Conditional Deep Learning (CDL) approach \cite{Conditional:journals/corr/PandaSR15}, our network leverages the difficulty of inputs to classify the relatively easy samples with early exits and conditionally activate the deeper layers. This technique is particularly suited for a distributed intelligence system where we can process the easy inputs at the energy-efficient edge device with quantized networks while only the inputs that are close to the decision boundary are sent to the cloud. As a result, both communication cost and computation cost can be improved. Through an extensive analysis of hybrid network architectures, we identify the tradeoffs in terms of energy, latency, accuracy as well as bit-precision of the layers for optimal hybrid network design. The contributions of our work are as follows:
\begin{itemize}
    \item We propose a methodology for designing hybrid precision neural networks for edge-cloud processing systems.
    \item We engineer early exiting strategies to conditionally activate the deeper layers of the neural networks for enforcing energy-efficiency in edge devices. 
    \item We evaluate the proposed methodology on state-of-art network architectures. We show that by modifying network configurations such as bit-precision at the edge, early exiting is triggered in most cases. 
\end{itemize}
The paper is organized as follows. The next section gives an overview of the design methodology for conditionally deep hybrid neural networks. Section III reports the experiment results and shows how to adjust network configurations to achieve a better performance-vs-energy trade-off.
\vspace{-0.2cm}

\section{Hybrid Network Design with Conditional Deep Learning}
\vspace{-0.2cm}
 In this section, we propose a methodology for designing hybrid neural networks and explore strategies to enable early classification in the edge network based on the difficulty of inputs to achieve early processing of data and reduction in energy consumption. 

\subsection{Hybrid Neural Networks}
We propose a hybrid neural network 
where a N-layer network is divided into two sections, $M$ layers where the bit-precision of the weights and inputs is $p_1$ and $N-M$ layers where the corresponding bit-precisions are $p_2$ where $p_2\gg p_1$. This is illustrated in Fig. \ref{fig_method}. Generally, $p_2$ can be considered as a 32-bit float as the primary objective for our design is to preserve the accuracy of the neural network. The design parameters $p_1$ and $M$ can be varied to identify the optimal division for the proposed hybrid networks considering the tradeoff between performance and energy consumption. 

\begin{figure}[h!]
\centering
\includegraphics[width=3.4 in]{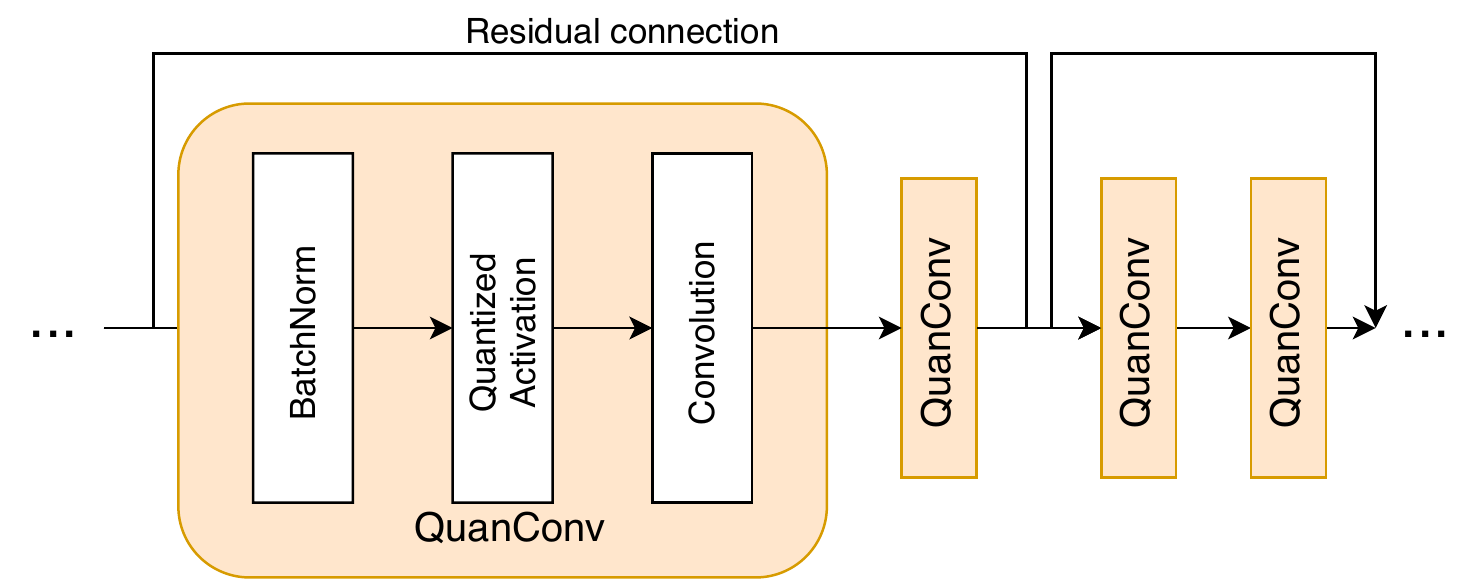}
\caption{Quantized convolutional layers in a ResNet}
\label{fig_res}
\end{figure}

The low-precision section of a hybrid CNN consists of QuanConv layers by which we quantize both the weights and inputs. Fig. \ref{fig_res} illustrates that a QuanConv layer is comprised of a batch normalization layer, an activation layer, and a convolution layer using quantized weights in order. In the case of a ResNet\cite{ResNet}, the residual connections are kept full-precision to mitigate the negative effect of quantized inputs with a small overhead. A batch normalization layer is applied before quantization to ensure that inputs hold zero mean\cite{XNOR:journals/corr/RastegariORF16}. Then the activation layer quantizes inputs using the following scheme. For an input matrix $I\in R^{c\times w_{i} \times h_{i}}$, where $c, w, h$ are the number channels, width and height respectively, the corresponding binary input matrix $I_B\in \{+1,-1\}^{c\times w_{i} \times h_{i}}$ and quantized input matrix $I_Q$ is given by 
\begin{align}
    I_B = \sign(I);\quad   I_Q = v_1 (\lfloor \dfrac{I+v_0}{v_1} \cdot Z \rceil /Z - v_2)
\end{align}
where $Z= 2^{p}-1$ is the number of quantization levels and $v_0, v_1, v_2$ are constants.

The quantization of weights is performed for each layer in a CNN. Let us represent the kernel as $W\in R^{c\times w \times h}$, where $c, w, h$ are the number channels and width and height of the kernel respectively. If both weights and inputs are extremely quantized (binarized), dot product operations in convolutions can be implemented efficiently using XNOR and bit-counting operations. 
To approximate the convolution operation, we estimate the real-value weight matrix $W$ using a quantized matrix $W_Q$ and a scaling factor $\alpha$ as proposed in \cite{XNOR:journals/corr/RastegariORF16}. 
\begin{align}
    I \ast W \approx (I_Q \ast W_Q)\alpha;\quad     \alpha = \dfrac{\|W\|_{l_1}}{n}\\
    W_Q= v_1 (\lfloor \dfrac{ W +v_0}{v_1} \cdot Z \rceil /Z - v_2)
\end{align}
where $n=c\times w \times h$ is a constant. The precision of weights and activations of the high precision section of a hybrid CNN are kept 32-bit floating point.  We adopt the training algorithm proposed by Rastegari et al \cite{XNOR:journals/corr/RastegariORF16} to train the networks with quantization.

In a hybrid neural network, the edge network must send extracted features corresponding to all samples to the cloud for further processing, which takes communication energy and time. In the next subsection, we will introduce how to enable the edge to analyze easy data independently.

\subsection{Conditional Exit in Deep Hybrid Networks}
\begin{figure}[h!]
\centering
\includegraphics[trim=140 140 140 140,clip,width=3.3 in]{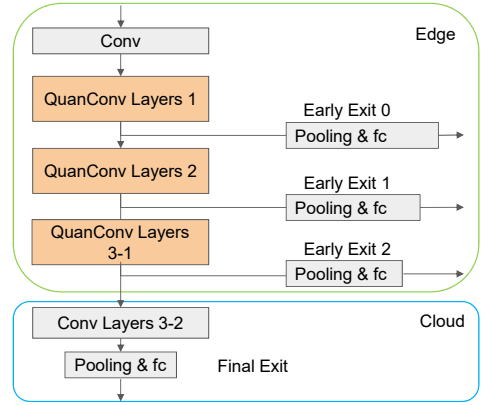}
\caption{An example of a hybrid neural network with three early exits. Blocks in orange are quantized. The sizes of early exits differ corresponding to their locations. }
\label{multiexit}
\end{figure}

Enabling distributed intelligence using hybrid networks in an AI-based fog computing system requires data processing both at the edge and the cloud. Since the difficulty of examples in a dataset often exhibits an inherent variability, we apply the concept of conditional deep learning \cite{Conditional:journals/corr/PandaSR15} on hybrid networks to enable classification of relatively simpler images at the low-precision edge, while the more difficult ones are passed to the high-precision deeper layers of the network in the cloud. Such conditional activation of deeper layers in the context of hybrid networks provides both communicational and computational energy efficiency and run-time improvements over standard networks. Each early exit at the edge is a linear classifier that consists of an average pooling layer and a fully connected layer. Typically, such an exit needs to be placed at the end of the low-precision section of the hybrid network, i.e, the part of the network being implemented on edge. However, more exits can be added in the low-precision section to facilitate early classification of easier examples. In Fig. \ref{multiexit}, we show an example of adding three early exits. The input to an early exit is the output from a convolutional layer, and all the early exits use full-precision weights to ensure no computation error occurs in the classification layer.
 
  

During the training process, although the early exits have already provided labels, the rest of the network would still be activated during training to optimize all exits based on the whole training set. Since there are more than one labels given by our network system, we need to modify the optimization objective to train both the hybrid network and additional exits. There are two optional training strategies: 1) Separate Optimization, 2) Joint optimization.

\subsubsection{Separate optimization}
In this approach, we first train the hybrid network without any early exits, hence, the same optimization objective can be used. Then we fix the parameters of all convolutional layers and train one early exit at a time. Let us denote the ground-truth label vector as $y$ and the output label given by early exit $k$ as $\hat{y}_k$. The optimization objective of the $k^{th}$ early exit is
\begin{align}
    L_k(\hat{y_k}, y) = -\sum_{c\in C} y_c \log \hat{y_k}_c
\end{align}
where $\hat{y_k}_c$ is a vector containing predicted probabilities for all possible class labels and $C$ represents the set of all classes. The backward propagation function will be called for $k$ times with respect to $L_k$. Since parameters of convolutional layers do not require gradient, the backward propagations in early exits are independent of each other.

\subsubsection{Joint optimization}
In this approach, we use a joint optimization technique to train the network along with early exits \cite{Branchy:journals/corr/abs-1709-01686}. The total loss is calculated as the weighted sum of the loss functions of each exit branch.
\begin{align}
    L(\hat{y},y) = \sum_{k=1}^K \lambda_k L_k(\hat{y}_{k}, y)
\end{align}
where $K$ is the total number of exits and $\lambda_k$ is set empirically and sum to 1. By joint optimization, the gradients used to update weights of the convolutional layers become
\begin{align}
    \dfrac{\partial L}{\partial w} = \sum_{k=1}^K \lambda_k \dfrac{\partial L_k}{\partial w}
\end{align}
Compared to separate optimization, training the entire network together optimizes the weights of the convolutional layers based on the losses of both the final exit and early exits, so it enables more samples to exit early and enlarges the energy saving. On the other hand, separate optimization is more flexible because it allows users to add early exits to a trained network. We will compare the results corresponding to these two optimization strategies in section III. Because our objective is to make as many inferences to finish at the edge as possible, we will use joint optimization for the rest of our experiments. 


\begin{algorithm}[h]
\SetAlgoLined
\KwIn{Samples $X$}
\KwOut{Classification results $\hat{Y}$}

 \For{sample i}{
    $O_1$[i] = Forward($X$[i],$W_t[1]$)\;
    \For{layer l = 2 to M}{ //\CommentSty{Propagate at the edge} \\
        $O_l$[i] = Forward($O_{l-1}$[i],$W_t[l]$)\;
        \If{Classifier k locates at layer l}{
            $\hat{Y}_k$[i] = Classifier[k]($O_l$[i])\;
            \If{Entropy($\hat{Y}_k$[i])$<Threshold_k$}{
                $\hat{Y}$[i] = $\hat{Y}_k$[i]\;
                break; //\CommentSty{take early exit}
            }
        }
    }
    \If{$\hat{Y}$[i] not found}{
        $\hat{Y}[i]$ = Forward($O_M$[i], $W_t[M:N]$);//\CommentSty{Continue propagating on the cloud} \\
    }
 }
 \caption{Inference using a conditionally deep hybrid neural network}
\end{algorithm}

Once the hybrid deep network is trained, we can utilize quantized weights and early exits to allow inference of easy samples to finish at the edge. 
Algorithm 1 shows the inference process. To classify a sample, we activate the layers of the deep network conditionally based on the entropy of the prediction result calculated by
\begin{align*}
    entropy(\hat{y}) = -\sum_{c\in C} \hat{y}_c \log \hat{y}_c
\end{align*}
 If the entropy is lower than the threshold, it means the early exit has high confidence about correctly labeling this sample, so the prediction can be returned without activating later layers. The thresholds for early exits are set empirically. 

\subsection{Design Considerations of Conditionally Deep Hybrid Network}
\subsubsection{Effect of bit-precision}
When conditional deep learning is applied to a hybrid network, the effect of bit-depth on the total energy consumption of the edge-cloud system becomes twofold. Although the energy consumed by an operation is proportional to the bit-depth, increasing the bit-depth means having a more precise network at the edge, so the number of activated high-precision operations would be fewer thanks to early exiting. As a result, using more bits for quantization does not necessarily lead to higher energy consumption. To achieve better performance-vs-energy trade-off, we will adjust the bit-depth depending on the complexity of datasets.

\subsubsection{Effect of the number of layers on edge}
In a conditionally deep hybrid network, the inference accuracy is not commensurate with the proportion of quantized layers due to additional exits. For simplicity, suppose that we have only one early exit located after the last layer at the edge. The prediction ability of the final exit on the cloud would be weakened to the degree corresponding to the number of quantized layers. However, the overall accuracy of the hybrid network depends on the accuracy of all exits, and that of the early exit is affected by the number of layers before it and the number of parameters contained in its fully-connected layer. As the number of quantized layers increases, the early exit can infer based on the high-level features extracted by a deeper network, which makes the inference task easier and potentially allows more samples to exit. Nevertheless, convolutional layers in different divisions of a ResNet have three distinct output sizes, and hence the number of parameters contained in the early exit drops significantly when moving to the later division. Therefore, it is important to find the optimal division of a hybrid network.

\subsubsection{Effect of the number of exits}
Using multiple exits at the edge can allow inference to terminate as soon as possible, thus the latency can be further reduced. In spite of that, whether it can also provide further energy saving to the edge depends on whether the power consumption of an earlier exit is smaller than the total power of convolutional layers and the later exit located in the next division of the network. Because convolutional layers are quantized but exits are not, their power consumption becomes a considerable part. If the number of classes in the dataset is large and the network is not very deep, using more than one early exits is likely to consume more power and hence is not desired.


\section{Experiments}
To validate the effectiveness of our model, we conduct several sets of experiments using PyTorch. We report results of ResNet32 with varying hybrid configurations and conditional deep learning over CIFAR10 and CIFAR100 datasets to explore the design space and evaluate the scalability. We estimate the total energy consumption of inference by multiplying the number of full-precision and binary operations with corresponding energy listed in Table. \ref{tab:energy}. Convolutions of matrices are implemented as multiply-and-add (MAC). For all our analysis, we have not included the communication energy between the edge and the cloud.

\begin{table}[h]
    \centering
\begin{tabular}{ |c|c||c|c| } 
 \hline
Operation & Energy (pJ) & Operation & Energy (pJ)\\
 \hline
32-bit Addition & 0.9 & Binary MAC & 0.2\\ 
 \hline
32-bit MAC & 4.6 & Memory access per bit &  2.5\\ 
 \hline
\end{tabular}
\\[8pt]
    \caption{Energy consumption chart}
    \label{tab:energy}
\end{table}

\subsection{Hybrid Network}
\begin{figure}[h!]
    \centering
    \includegraphics[width=2.5 in]{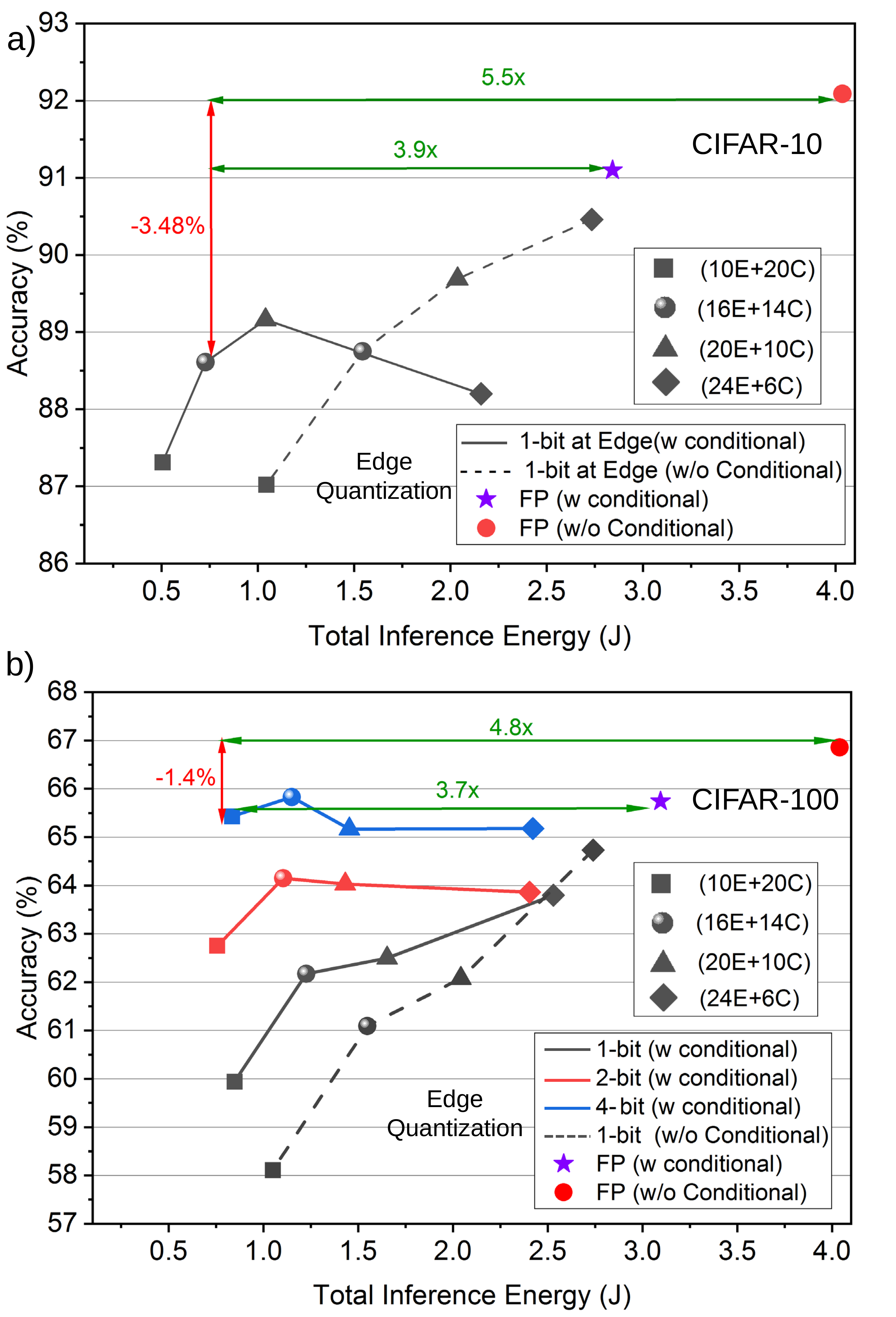}
     \vspace{-0.3cm}
    \caption{a) Accuracy vs Energy plot for CIFAR-10 showing 20E+10C config achieves 5.5x energy benefit over a full-precision standalone network on cloud. b)Accuracy vs Energy plot for CIFAR-100 showing 4-bit network at edge (24E+6C config) achieves 4.8x energy benefit over a full-precision standalone network on cloud. Various configurations have been explored with different edge and cloud layers.}
    \label{fig_cifar100}
    \vspace{-0.5cm}
\end{figure}

We present the inference accuracy and energy results for hybrid networks in
Fig. \ref{fig_cifar100} using dashed lines. ``10E + 20C'' represents a hybrid network with 10 quantized layers at the edge and 20 layers at the cloud. When we binarize the first 10 layers of the network, the energy reduction with respect to a full-precision ResNet-32 is $1.5\times$ for both CIFAR10 and CIFAR100 with accuracy losses of $1.5\%$ and $2.1\%$ respectively. On binarization of 20 layers, the energy saving becomes $2.6\times$.

\subsection{Conditionally Deep Hybrid Network}
We apply conditional deep learning on hybrid networks by adding an early exit after the last quantized layer at the edge. The thresholds for entropy to determine whether inference can finish at the edge are set to 0.5 for CIFAR10 and 1.0 for CIFAR100. In Fig. \ref{fig_cifar100}(a), we show the energy consumption and accuracy on CIFAR10 with and without an early exit. The energy saving of binarizing 20 convolutional layers becomes $5.5\times$ after adding an early exit, which is two times larger than before, and the accuracy remains the same. In comparison with a full-precision network with conditional exiting, it is 3.9$\times$ more energy efficient due to hybrid quantization. Next, we evaluate the different configurations of the conditional deep hybrid network as mentioned in Section II.
\subsubsection{Separate Optimization VS Joint Optimization}
In Table. \ref{training_compare}, we compare the results corresponding to these two different optimization strategies. For joint optimization, we set the weights $\lambda_k$ used to calculate the joint loss empirically to $0.6$ for the early exit because giving more weight to the early exit will encourage more discriminative feature learning in early layers\cite{Branchy:journals/corr/abs-1709-01686}. With the same hybrid network configuration (16 E + 14 C), joint optimization enables more samples to finish inference at the early exit and yields higher accuracy. It confirms our assumption that using joint optimization can make the edge network more confident and make the edge-cloud system more energy efficient. 

\begin{table}[h]
    \centering

\begin{tabular}{ |c|c|c|c|c| } 
\hline
\multirow{2}{*}{}& \multicolumn{2}{c|}{Percentage of early exiting (\%)} & \multicolumn{2}{c|}{Accuracy(\%) } \\
 \cline{2-5}
 & Separate & Joint & Separate & Joint\\
  \hline
 CIFAR10 & 24.1 &  52.8 & 88.72 & 89.16  \\ 
 \hline
 CIFAR100 & 17.3 & 24.6  & 61.80 &  62.50\\ 
 \hline
 \end{tabular}
\\[8pt]
    \caption{Comparison between separate optimization and joint optimization}
    \label{training_compare}
\end{table}

\subsubsection{Effects of hybrid bit-depth}
Unlike CIFAR10 of which most samples can be confidently inferred by the binarized network at the edge, CIFAR100 is more complex so the binarized network is only confident on less than 30\% of test samples as illustrated in Fig. \ref{fig_percent}. To make the network at the edge more confident, we use 2-bit or 4-bit quantization instead of binarization. For each quantization option, we vary the number of quantized layers and plot the performance of conditionally deep hybrid networks in Fig. \ref{fig_cifar100}(b). Points in the top-left corner are corresponding to optimal networks since they achieve high accuracy and low inference energy. The experiment results show that quantizing the network with more bits can achieve better accuracy without trading off energy efficiency. 4-bit and 2-bit quantization only cause $1.4\%$ and $3.0\%$ accuracy degradation respectively, while energy savings can be $4.8\times$ and $5.3\times$ compared to a full-precision network. In Table. \ref{tab:cifar100-edge}, we show the inference energy consumed by the edge and the cloud. From the top left corner to the bottom right corner, the cloud energy decreases rapidly while the edge energy does not change much because we keep the first layer at the edge full-precision and it dominates the edge energy. 

\begin{table}[h]
    \centering
\begin{tabular}{ |c|c|c|c|c|c|c| } 
 \hline
\multirow{2}{*}{Energy (mJ)} &  \multicolumn{2}{c|}{1-bit} & \multicolumn{2}{c|}{2-bit} & \multicolumn{2}{c|}{4-bit} \\
 \cline{2-7}
 & edge & cloud & edge & cloud & edge & cloud \\
 \hline
10 E + 20 C & 261 & 2269 &	295 & 2108 & 363 & 2059\\ 
 \hline
16 E + 14 C & 213 & 1440 &	264& 1168 &	365 & 1089 	\\ 
 \hline
20 E + 10 C & 236 & 991 &	299 & 805 &	424 & 726	\\ 
 \hline
24 E + 6 C & 203 & 645	& 278 & 478  &	429 & 406
	\\ 
 \hline
\end{tabular}
\\[8pt]
    \caption{Inference energy at the edge and cloud}
    \label{tab:cifar100-edge}
    \vspace{-0.4cm}
\end{table}

\subsubsection{Effect of the number of layers on edge}
\begin{figure}[t]
\centering
\includegraphics[width=2.5 in]{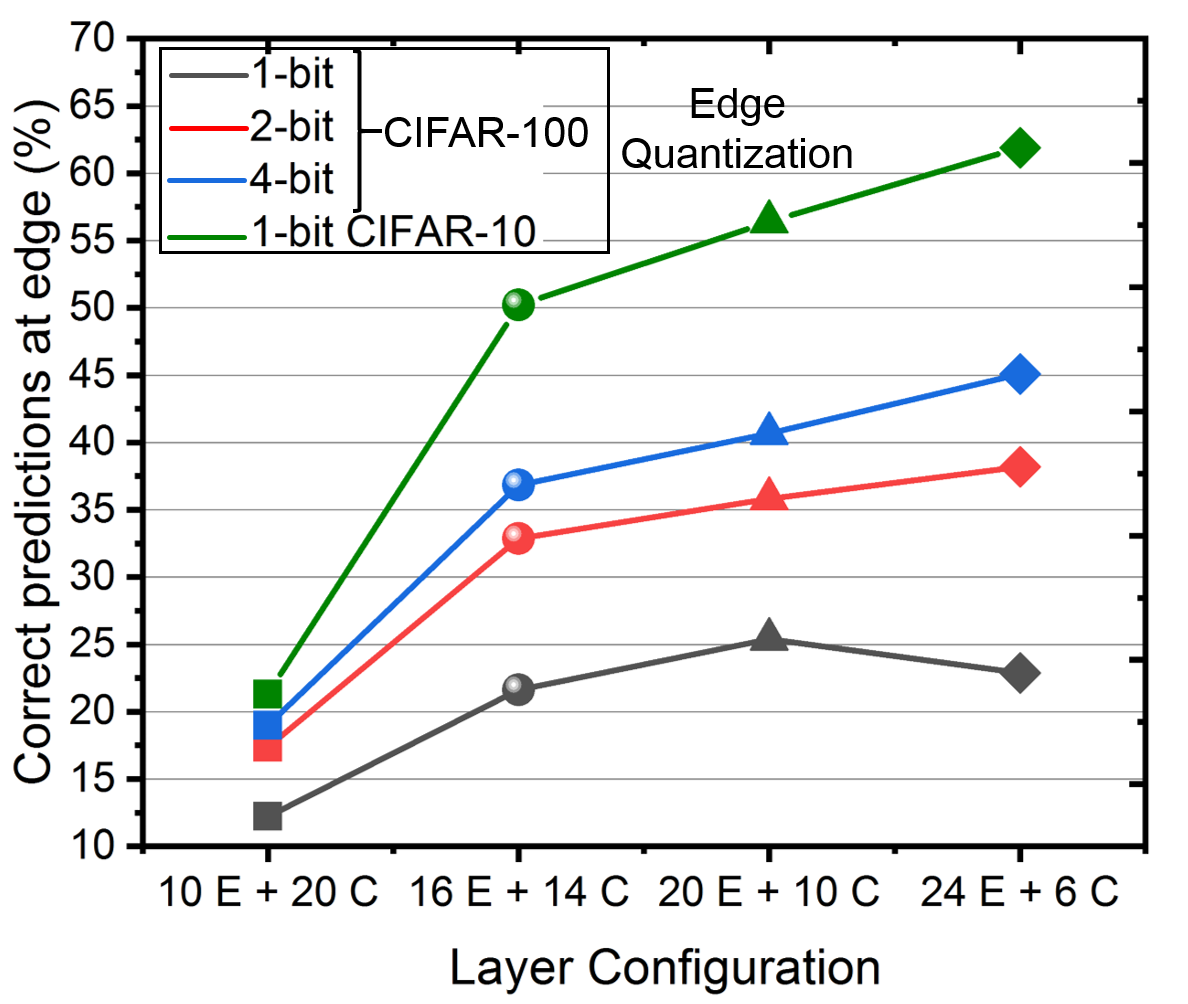}
\vspace{-0.3cm}
\caption{Accuracy and Percentage of early exiting samples at the edge for CIFAR-100 dataset on Resnet-32}
\label{fig_percent}
\end{figure}

 From Fig. \ref{fig_cifar100}(b), we can see that as the number of binary layers increases, the hybrid network becomes less accurate. However, this is not true if early exits are added, as we have discussed in Section II.C. In Fig. 4, we find the sweet spots which provides the best performance is using 16 binary layers and 14 full-precision ones. We also have discussed that the precision of the edge network depends on the number of parameters contained in the early exit. In Fig. \ref{fig_percent}, we show that in most cases, the number of correct predictions finish at the edge increases with the number of quantized layers, however, when the early exit is located after 24 binary layers, its weight matrix of size $256\times 100$ is not robust enough, and hence the number of early exiting samples on CIFAR100 abnormally decreases. In the case of 2-bit or 4-bit quantization, the network is more robust, so the effect of the decrease in the number of parameters is not as severe as that on binarized networks.

\subsubsection{Effects of the number of exits}

\begin{table}[h]
    \centering
\begin{tabular}{ |c|c|c|c|c|c|c| } 
 \hline
& \multicolumn{2}{c|}{\% of exiting at edge} & \multicolumn{2}{c|}{Accuracy} & \multicolumn{2}{c|}{Energy(mJ)}\\
 \cline{2-7}
\# of early exits & 1 & 2 & 1 & 2  & 1 & 2  \\
 \hline
20 E + 10 C & 59.3 & 38.6, 19.7 & 88.61 & 85.98 & 729 & 724\\ 
 \hline
\end{tabular}
\\[8pt]
    \caption{Effect of the number of early exits on CIFAR10}
    \label{multi}
    \vspace{-0.3cm}
\end{table}
In Table. \ref{multi}, we compare the results of hybrid networks with one or two early exits on CIFAR10. We evenly distribute two exits: one is at the end of the edge, the other is put in the middle. We notice that adding more exits has relatively minor effects on the total energy. This is because the full-precision layers of the network are much more power-hungry than the binarized ones, and the total energy of inference is dominated by the energy of samples that cannot be handled at the edge. As illustrated in Table. \ref{multi}, adding more exits does not enable more samples to skip the full-precision layers, so the cloud energy is not reduced. Besides, using two early exits causes performance degradation due to the complexity of training. For CIFAR100, since the number of classes is 10 times larger, adding the other early exit in Conv 1 would lead to even higher energy consumption than activating binarized Conv 2 and the early exit in Conv 2. Moreover, if two exits are close, the number of exiting samples at the later one would be small because their confidences are similar. Therefore, using one early exit at the edge is enough in most cases. 

\section{Conclusion}
In conclusion, we applied conditional deep learning on partially quantized neural networks to simulate neural network systems distributed at the edge and cloud. We found that the joint optimization strategy can enable the edge to handle more samples. We showed that conditionally deep hybrid networks achieve much better performance-vs-energy trade-offs than quantized networks without early exits. By modifying the bit-depth and number of quantized layers at the edge, we found the optimal configurations for hybrid networks, which allow 65\% of CIFAR10 samples and 52\% of CIFAR100 samples to exit early and achieve around five times energy reduction.


\ifCLASSOPTIONcaptionsoff
  \newpage
\fi

\bibliographystyle{ieeetr}
\bibliography{ref}

\end{document}